\documentclass[lettersize,journal]{IEEEtran}
\usepackage{amsmath,amsfonts}
\usepackage{algorithmic}
\usepackage{algorithm}
\usepackage{array}
\usepackage[caption=false,font=normalsize,labelfont=sf,textfont=sf]{subfig}
\usepackage{textcomp}
\usepackage{stfloats}
\usepackage{url}
\usepackage{verbatim}
\usepackage{graphicx}
\usepackage{cite}
\usepackage{xspace}
\usepackage{optidef}
\usepackage{makecell}
\usepackage{multirow}
\usepackage{epsfig}
\usepackage{caption}
\usepackage{subfig}

\begin{document}

\title{Leveraging unsupervised data and domain adaptation for deep regression in low-cost sensor calibration}

\author{
Swapnil Dey$^*$, Vipul Arora$^*$, Sachchida Nand Tripathi
\thanks{Swapnil Dey and Vipul Arora are with the Department of Electrical Engineering, Indian Institute of Technology Kanpur, Kanpur 208016, India (e-mail: swapon@iitk.ac.in, vipular@iitk.ac.in).

Sachchida Nand Tripathi is with the Department of Civil Engineering, Indian Institute of Technology Kanpur, and the Centre for Environmental Science and Engineering, Indian Institute of Technology Kanpur, Kanpur 208016, India (e-mail: snt@iitk.ac.in).

$^*$ Equal contribution}
}


\maketitle

\begin{abstract}
Air quality monitoring is becoming an essential task with rising awareness about air quality. Low cost air quality sensors are easy to deploy but are not as reliable as the costly and bulky reference monitors. The low quality sensors can be calibrated against the reference monitors with the help of deep learning. In this paper, we translate the task of sensor calibration into a semi-supervised domain adaptation problem and propose a novel solution for the same. The problem is challenging because it is a regression problem with covariate shift and label gap. We use histogram loss instead of mean squared or mean absolute error, which is commonly used for regression, and find it useful against covariate shift. To handle the label gap, we propose weighting of samples for adversarial entropy optimization. In experimental evaluations, the proposed scheme outperforms many competitive baselines, which are based on semi-supervised and supervised domain adaptation, in terms of R$^2$ score and mean absolute error. Ablation studies show the relevance of each proposed component in the entire scheme.
\end{abstract}


\section{Introduction}
Deep learning based models achieve remarkable performances with labeled data. However, many practical scenarios face a scarcity of labeled data, while there is an abundance of unlabeled data. Semi-supervised approaches have been developed to make use of the large unlabeled data in those scenarios.
Another challenge faced in real world tasks is the mismatch in the data distributions across domains. The models trained in one domain, called source domain, are unable to generalize well to the other, called target domain. Domain adaptation \cite{ben2006analysis} based approaches come to rescue here. 
While there have been several works on semi-supervised domain adaptation for classification \cite{yao2015semi, singh2021clda, daume2010frustratingly, kumar2010co, he2020classification}, only a limited number of works address this challenge for regression.
In this paper, we focus on a regression problem where large amounts of labeled data is available in the source domain, while a very limited labeled data, along with a large amount of unlabeled data, is available in the target domain. We apply our approach to calibration of low cost PM$_{2.5}$ sensors for air quality monitoring.

Saito \emph{et al.} \cite{saito2019semi} use Minmax Entropy (MME) based approach for semi-supervised domain adaptation in classification problems. The goal there is to not just align the feature distribution in target domain with that in the source domain, but also to learn better class-discriminative boundaries in the target domain. They achieve this by minimizing the entropy with respect to the feature extractor and maximizing the same with respect to the classifier. In this paper, we extend their method to regression. 

Since, regression does not include discrete classes that is needed for the MME approach, we convert the continuous target into a probability mass function by binning.
Apart from allowing semi-supervised domain adaptation, the histogram based regression brings in other benefits too. Imani and White \cite{imani2018improving} find that histogram based loss function improves regression performance by regularizing the learning. In this paper, we amend the MME based approach for regression problems.

Label sets of large amount of source and limited number of target domain inputs could be different from that of the unlabeled target domain inputs. We compare this with having novel classes in the unlabeled target domain inputs, not seen in the labeled source and target domain inputs. These novel classes could have a negative effect on learning. Peng \emph{et al.} \cite{peng2019investigating} propose a distance based weighting framework for semi-supervised learning in classification. We adapt the same framework for domain adaptation in the regression task at hand.

Air pollution monitoring is generally done using costly high-fidelity monitors \cite{sahu2020validation}. These days, with increased awareness about air quality, the need of low cost solutions is being felt. Low cost sensor devices (LCSD) are affordable and portable but offer less reliable measurements \cite{sahu2021robust}, \cite{snyder2013changing}. To improve the precision and robustness of LCSD, deep calibration methods have been found to be effective. The supervision for these deep calibration models comes by co-locating the LCSD with the costly reference monitors, which is cumbersome. Hence, there is a scarcity of labeled data, while an abundance of unlabeled data. There are various factors such as the environmental conditions, geographical location and sensor characteristics, that bring a mismatch in domains and make domain adaptation a necessity. We apply the proposed semi-supervised domain adaptation method to sensor calibration problem.

The main contributions of this paper are:
\begin{itemize}
    \item a novel semi-supervised domain adaptation algorithm for regression tasks
    \item a weighted min-max entropy loss for learning from the unlabeled data by giving more weight to the unlabeled samples that are closer to the labeled samples
    \item a novel algorithm for sensor calibration that can leverage even the unlabeled sensor data at the target domain. To the best of our knowledge, semi-supervised deep learning has not been used for sensor calibration before.
\end{itemize}

The data and the codes to implement the work described in this paper are available at \url{https://github.com/madhavlab/2022_SSDA_airquality/}.

\section{Related Works}

\subsection{Domain Adaptation}
Generally, domain adaptation problem is characterized by a source distribution, $S(x,y)$ and a target distribution, $T(x,y)$, where $x$ is the input and $y$ are the corresponding labels. Domain adaptation problem is characterized by co-variate shift where the marginal distributions of two corresponding domain differ, i.e., $S(x)\neq T(x)$. To minimise this co-variance shift, several feature-based approaches have been proposed where the model learns a transformation for the feature space to correct the domain shift between $S(x)$ and $T(x)$. In \cite{tzeng2014deep}, the authors propose a new adaptation layer in model that learns to forget the distributional discrepancy between the two domains by minimizing the distributional distance (Maximum Mean Discrepancy). In \cite{long2018transferable}, authors propose Deep Adaptation Network (DAN), where the mean embeddings of representations from different domains are matched in reproducing kernel Hilbert space (RKHS).
In \cite{sun2016deep} and \cite{sun2016return}, authors propose correlation alignment based loss to minimize the domain discrepancy. In the past few years, there have been extensive research to develop adversarial methods for domain adaptation \cite{ganin2016domain, shen2018wasserstein, richard2020unsupervised, tzeng2017adversarial, zhang2019bridging, long2018conditional}. In \cite{ganin2016domain}, the authors propose learning deep features which are discriminative for the task at hand, whilst invariant with respect to the shift between domains. In \cite{richard2020unsupervised}, authors propose to minimize hypothesis discrepancy between multiple source domains and target domain to make the learned representations to be domain invariant. The work \cite{tzeng2017adversarial} extends the method of \cite{ganin2016domain} by proposing a two-stage algorithm which first learn source encoder and task hypothesis from the labeled source data, and later, learns a target encoder through adversarial training. In \cite{zhang2019bridging}, authors suggest to learn a new domain-invariant representation by minimizing the margin discrepancy distance between encoded source and target domains. In \cite{long2018conditional}, model also learns domain-invariant features through multi-linear conditional adversarial training between feature extractor and domain classifier. There are other domain adaptation methods \cite{saito2018maximum, saito2017adversarial, saito2019semi} which not only align feature distributions from different domains, but also concentrate to learn better class discriminative boundaries. We adapt the idea of \cite{saito2019semi} for the regression task presented in this paper. 

\subsection{Semi-supervised Learning}
Supervised machine learning has had fabulous success in various applications. However, it needs labeled data for training the models.
There are many applications where unlabeled data is available in plenty but labeled data is not easy to obtain. Semi-supervised learning methods \cite{4724730, yang2021survey, 9737635} have been developed to leverage the unlabeled data by learning good representations from it and subsequently mapping it to the target labels by learning from the labeled data.
Contrastive learning \cite{9817089} is one of the most popular ways to learn from unlabeled data. Semi-supervised learning is also used for enhancing domain adaptation \cite{6774431, saito2018maximum, saito2019semi}.

One of the problems with limited label data is class imbalance, where not all classes may be represented in a balanced way in the labeled data. In supervised learning, several approaches have been proposed \cite{9656689, Arora2019}. For semi-supervised learning too, there are some approaches for handling class imbalance \cite{9512275, kim2020distribution}. These approaches are mostly designed for classification tasks. For regression tasks, the imbalance problem becomes more challenging as there is no finite number of classes in regression.

\subsection{Sensor Calibration}

Dense deployment of Continuous ambient air quality monitoring stations (CAAQMS) can provide highly reliable real-time PM$_{2.5}$ values with having the shortcoming of their high cost \cite{sahu2020validation}. However, small, portable low-cost sensor devices (LCSD) can be deployed densely compromising the reliability of their measurements \cite{snyder2013changing}, \cite{sahu2021robust}. Compromise in the reliability of measured values by low-cost sensors raise the need of calibration of LCSD against CAAQMS\cite{zheng2018field}.

Extensive research has been carried out in low-cost sensor calibration for the past few years \cite{delaine2019situ},\cite{wang2020category}. In \cite{zheng2019gaussian}, they propose a linear regressor and Gaussian process regressor for the calibration of low-cost PM$_{2.5}$ sensors. The work \cite{zheng2018field} find quadratic calibration model to be better than its linear counterpart. Statistical methods, such as ARIMA based models have been also suggested to calibrate low-cost sensors in \cite{munir2021application}. The work \cite{sahu2021robust} proposes Mahalanobis distance based weighted K-nearest neighbour algorithm with a learned metric for calibration. Neural network based methods such as fully connected neural network \cite{spinelle2017field},\cite{spinelle2015field}, convolutional neural network (CNN) \cite{bagkis2021analyzing}, Recurrent neural network \cite{han2021calibrations} have also been found to be effective in achieving state-of-the-art results for sensor calibration. But none of these works leverage domain adaptation for calibration.

There are two recent works \cite{jha2021domain},\cite{yadav2021few} which involve domain adaptation for sensor calibration. \cite{jha2021domain} applies simple fine tuning based domain adaptation technique for calibration. In \cite{yadav2021few}, authors propose to apply model-agnostic meta learning technique for few shot domain adaptation based calibration. But none of these domain adaptation based calibration methods leverages unlabeled data from the target domain that is available in abundance.

\section{Proposed Method}
In semi-supervised domain adaptation, we are given labeled data and their corresponding ground truth from the source domain $\mathcal{D}_{s,l}$ = \{${(\mathbf{x}_{i}^{s,l},y_{i}^{s,l})}\}_{i=1}^{m_s}$. From the target domain, we are given a few data-label pairs $\mathcal{D}_{t,l}$ = \{${(\mathbf{x}_{i}^{t,l},y_{i}^{t,l})}\}_{i=1}^{m_t}$ and a large amount of unlabeled data $\mathcal{D}_{t,u}$ = \{${(\mathbf{x}_{i}^{t,u})}\}_{i=1}^{m_u}$ where $m_s>>m_t$.


\subsection{Histogram Loss}

Commonly, linear regression involves minimizing L2 loss or squared error loss. This can be interpreted probabilistically as maximum likelihood estimation of output (conditioned on input) modeled as a Gaussian random variable with a fixed variance. The final prediction is the mean of this distribution.
Histogram loss \cite{imani2018improving} involves generalizing the model by giving up the Gaussian assumption of the model and by computing the density function instead of just a point estimate.

Let the input be $\mathbf{x}\in\mathbb{R}^{d}$ and the output be a continuous random variable, conditioned on the input, as $y\sim p(y|\mathbf{x})$. 
$p(y|\mathbf{x})$ is modeled as a multinomial distribution, or a histogram, $p(\tilde{y}|\mathbf{x})$, by discretizing the continuous target output $y$ to $\tilde{y}$. $K\in \mathbb{Z}$ discrete bins are created by partitioning the support range of $y$ uniformly. 
While $p(\tilde{y}|\mathbf{x})$ is the target distribution, the estimated distribution is $q(\tilde{y}|\mathbf{x})$. Finally, the point-estimate is 
\begin{align}
    \hat{y}=\mathbb{E}_{\tilde{y}\sim q(\tilde{y}|\mathbf{x})}[\tilde{y}].
\end{align}
To estimate $q(y|\mathbf{x})$, the KL divergence between $q$ and $p$ can be minimized.
\begin{align*}
    q &= \arg\min_q KL[p||q] \\
    &= \arg\min_q H[p,q]-H[p] \\
    &= \arg\min_q H[p,q]
\end{align*}

Here, $H[p,q]$ is the cross-entropy between target and estimated distributions, respectively, in the dataset $\mathcal{D}$.
For the multinomial distributions $p$ and $q$, the cross-entropy term is given by
\begin{align}
    H_\mathcal{D}[p,q] = -\mathbb{E}_{(\mathbf{x},y)\sim \mathcal{D}} \big[\sum_{k=1}^K p(\tilde{y}[k]|\mathbf{x}) \log(q(\tilde{y}[k]|\mathbf{x}))\big]
\end{align}
This can be approximated from the given training data using Monte Carlo approximation as
\begin{align}
    H_\mathcal{D}[p,q] = -\frac{1}{M}\sum_{j=1}^M\sum_{k=1}^K p(\tilde{y}[k]|\mathbf{x}_j) \log(q(\tilde{y}[k]|\mathbf{x}_j)). \label{eq:histloss}
\end{align}
Here, $M$ is the number of training data samples.

Eq.~\eqref{eq:histloss} provides a supervised loss term for the regression problem at hand. This term can be used for the labeled data, viz., $\mathcal{D}_{s,l}$ and $\mathcal{D}_{t,l}$.

\subsection{Min-Max entropy loss}
For semi-supervised domain adaptation for classification, \cite{saito2019semi} uses min-max entropy formulation. 
The goal of this formulation is to make use of the unlabeled data from the target domain to improve the model performance in the target domain.
The model comprises a feature encoder $F_\theta$ and a classifier $F_\phi$.
The central idea is to maximize the entropy of the entropy of predictions over $\mathcal{D}_{t,u}$ with respect to $\phi$ and to minimize the same with respect to $\theta$. The two optimization steps are repeated as in a mini-max game over $\mathcal{D}_{t,u}$. This helps the feature extractor $F_\theta$ to generalize over the unlabeled target data by reducing the bias towards the source data. The classifier $F_\phi$ outputs do become more uncertain as the loss term aims at increasing the variance but the labeled target data $\mathcal{D}_{t,l}$ holds them from degradation via the supervised loss term. 

We define our model as consisting of two parts, namely, a feature encoder $F_\theta$ and a histogram estimator $F_\phi$, so that 
\begin{align}
    q(\tilde{y}|\mathbf{x}) = F_\phi(F_\theta(\mathbf{x})).
\end{align}
The entropy of output is given by 
\begin{align}
H_{\mathcal{D}_{t,u}}[q]&=-\mathbb{E}_{\mathbf{x}\in\mathcal{D}_{t,u}}\big[\sum_{k=1}^K q(\tilde{y}[k]|\mathbf{x})\log q(\tilde{y}[k]|\mathbf{x})\big] \nonumber \\
&=-\frac{1}{|\mathcal{D}_{t,u}|}\sum_{j=1}^{|\mathcal{D}_{t,u}|}\sum_{k=1}^{K} q(\tilde{y}[k]|\mathbf{x}_j) \log q(\tilde{y}[k]|\mathbf{x}_j) \label{eq:entloss}
\end{align}
where $|\mathcal{D}_{t,u}|$ is the number of samples in the target unlabeled data.

\subsection{Weighted Min-Max Entropy loss}
Minmax entropy loss handles the problem of domain gap, a.k.a. covariate shift, i.e., the difference in the distributions of $\mathbf{x}$ in the source and target domains. However, there is another problem, called label gap, where $\mathcal{D}_{t,u}$ may not contain similar labels as those of $\mathcal{D}_{s,l}$ and $\mathcal{D}_{t,l}$. \cite{litrico2021semi} use label normalization to handle label gap. In this paper, we propose sample weighting for the same problem. The idea of sample weighting is also used by \cite{peng2019investigating} for extreme classification problems.
We down-weight the loss term of Eq.~\eqref{eq:entloss} for those $\mathbf{x}_j$ which may correspond to labels far from the labels present in the labeled datasets. We measure this distance in the encoded feature space. 

Given an $\mathbf{x}_j\in \mathcal{D}_{t,u}$, a score $S_j$ is assigned to it as
\begin{align}
    S_j = e^{-\beta d_{j}}
\end{align}
where $\beta\in\mathbb{R}^+$ is a hyperparameter to control the decay of exponential function and 
\begin{align}
    d_j = \min_{\mathbf{x}_{j'}\in \mathcal{D}_{s,l} \cup \mathcal{D}_{t,l}} ||F_\theta(\mathbf{x}_j)-F_\theta(\mathbf{x}_{j'})||.
\end{align}
The weighted min-max entropy loss then takes the form
\begin{align}
    H_{\mathcal{D}_{t,u}}'[q] = -\frac{1}{|\mathcal{D}_{t,u}|}\sum_{j=1}^{|\mathcal{D}_{t,u}|}\sum_{k=1}^{K} S_j q(\tilde{y}[k]|\mathbf{x}_j) \log_2 q(\tilde{y}[k]|\mathbf{x}_j). \label{eq:wentloss}
\end{align}

\subsection{Final training objective}
The final loss function is given by
\begin{align}
    \mathcal{L} = H_{\mathcal{D}_{s,l}}[p,q] + H_{\mathcal{D}_{t,l}}[p,q] + \alpha(t) H_{\mathcal{D}_{t,u}}'[q]
\end{align}
where, $\alpha(t)\in\mathbb{R}^+$ is a weighting hyperparameter and a function of epoch index $t$. 

The magnitude of $\alpha(t)$ controls the relative weights of the supervised and unsupervised losses. The parameters are optimized as
\begin{align}
    \theta &= \arg\min_{\theta} \mathcal{L}; \alpha(t)=|\alpha(t)| \\
    \phi &= \arg\min_{\phi} \mathcal{L}; \alpha(t)=-|\alpha(t)|
\end{align}
The opposite signs of $\alpha(t)$ in the above equations are due to the min-max entropy loss. The scheduling of $|\alpha(t)|$ is important in iterative training. If $|\alpha(t)|$ is set as high in the beginning of training, it will lead to poor training because the model estimates are not reliable, and hence the unsupervised loss term harms more than helping. Hence, we start with a small value of $|\alpha(t)|$ and increase it gradually until it reaches a constant value.
\begin{align*}
|\alpha(t)| = \begin{cases} 0, & t\leq T_{1} \\
\frac{(t-T_1)}{(T_2-T_1)}, & T_1<t \leq T_2 \\
|\alpha|_\infty, & t>T_2
\end{cases}
\end{align*}
where $|\alpha|_\infty$, $T_1$ and $T_2$ are set heuristically. 
To implement the min-max part, we use a gradient reversal layer to flip the gradients for $\theta$ and $\phi$.

\section{Experiments}
\subsection{Dataset description, pre-processing and feature selection}
We use a dataset \cite{jha2021domain} of co-located LCSDs from 11 different locations around Mumbai city. One location is taken as the source location and the rest 10 as the target location. The low-cost sensors give $4$ measurement signals namely PM$_{2.5}$, PM$_{10}$, Temperature and Humidity, with a sampling rate of 1 per hour.
The co-located reference monitors give PM$_{2.5}$ values, with the same sampling rate. 
As in \cite{jha2021domain}, in addition to the raw features (samples) from LCSD, we use a set of 23 derived features for training the calibration model. In \cite{jha2021domain}, authors find a strong correlation of these $27$ features with reference PM$_{2.5}$ values, and report a better model performance with $27$ features as compared to that with $4$ raw features. 
The dataset is pre-processed to remove the outliers and missing values as done by \cite{jha2021domain}.
Before training, we also perform feature standardization \cite{zheng2018feature} of input data to reduce the sensitivity of model to different ranges of different features during training.

For our experiments, we use 52 days of data as training set, 14 days of data as validation set and 14 days of data as test set at the source location. The data is not randomized because these are time series with samples correlated across time.
At the target location, we use only $2$ days of data as the labeled training set.
The data durations at the target locations are shown in Table~\ref{tab:data}.
\begin{table}[h!]
  \begin{center}
    \caption{Data durations for unlabeled training, validation and test sets at various target locations. Labeled training set is of 2 hours duration at each location.}
    \label{tab:data}
    \begin{tabular}{|c|l|l|l|} 
    \hline
      \textbf{target loc.} & \textbf{unlab. train}&\textbf{Validation}&\textbf{Testing}\\
      \hline
      loc 1 & 38 days & 7 days & 25 days\\
      \hline
      loc 2 & 38 days & 7 days & 25 days\\
      \hline
      loc 3 & 38 days & 7 days & 25 days\\
      \hline
      loc 4 & 38 days & 7 days & 16 days 15 hours\\
      \hline
      loc 5  & 38 days & 7 days & 25 days\\
      \hline
      loc 6 & 25 days & 7 days & 10 days 8 hours\\
      \hline
      loc 7 & 16 days 15 hours & 5 days & 10 days\\
      \hline
      loc 8 & 38 days & 7 days & 25 days\\
      \hline
      loc 9 & 22 days & 4 days 15 hours & 11 days 2 hours \\
      \hline
      loc 10 & 38 days & 7 days & 7 days 15 hours\\
      \hline
    \end{tabular}
  \end{center}
\end{table}
\begin{table*}[t]
\centering
\caption{R$^2$ score at different target locations for different models. Location indices are in columns and models are in rows. The higher the better.}
\label{table:r2_score}
\begin{tabular}{|p{2.1cm}||p{0.9cm}|p{0.9cm}|p{0.9cm}|p{0.9cm}|p{0.9cm}|p{0.9cm}|p{0.9cm}|p{0.9cm}|p{0.9cm}|p{0.9cm}|p{0.9cm}|}
 \hline
 \multicolumn{12}{|c|}{R$^2$ score}\\
 \hline
   \textbf{Model} & loc1 & loc2 & loc3 & loc4 & loc5 & loc6 & loc7 & loc8 & loc9 & loc10 & avg. \\ [0.5ex]
 \hline
 UNCAL&\makecell[l]{71}&\makecell[l]{$70$}&\makecell[l]{-224}&\makecell[l]{$66$}&\makecell[l]{-25}&\makecell[l]{5}&\makecell[l]{-1358}&\makecell[l]{$77$}&\makecell[l]{$4$}&\makecell[l]{51}&\makecell[l]{-126.3}\\
 \hline
 DANN&\makecell[l]{44}&\makecell[l]{$79$}&\makecell[l]{-103}&\makecell[l]{$86$}&\makecell[l]{50}&\makecell[l]{44}&\makecell[l]{-16}&\makecell[l]{$61$}&\makecell[l]{$46$}&\makecell[l]{2}&\makecell[l]{29.3}\\
 \hline
 ADDA&\makecell[l]{61}&\makecell[l]{-12}&\makecell[l]{10}&\makecell[l]{6}&\makecell[l]{57}&\makecell[l]{-114}&\makecell[l]{-196}&\makecell[l]{57}&\makecell[l]{-64}&\makecell[l]{-46}&\makecell[l]{-24.1}\\
  \hline
 DEEP-CORAL&\makecell[l]{$25$}&\makecell[l]{$82$}&\makecell[l]{-97}&\makecell[l]{$69$}&\makecell[l]{$34$}&\makecell[l]{$40$}&\makecell[l]{-36}&\makecell[l]{$38$}&\makecell[l]{$52$}&\makecell[l]{-1}&\makecell[l]{$20.06$}\\
   \hline
  DDC&\makecell[l]{43}&\makecell[l]{$78$}&\makecell[l]{-137}&\makecell[l]{$72$}&\makecell[l]{50}&\makecell[l]{-12}&\makecell[l]{-999}&\makecell[l]{$4$}&\makecell[l]{$28$}&\makecell[l]{-2}&\makecell[l]{-87.5}\\
  \hline
  MDD&\makecell[l]{39}&\makecell[l]{$82$}&\makecell[l]{-71}&\makecell[l]{$85$}&\makecell[l]{33}&\makecell[l]{$43$}&\makecell[l]{$5$}&\makecell[l]{$62$}&\makecell[l]{$52$}&\makecell[l]{-4}&\makecell[l]{$32.6$}\\
  \hline
 TLMAE&\makecell[l]{$61$}&\makecell[l]{$70$}&\makecell[l]{-11}&\makecell[l]{$81$}&\makecell[l]{$72$}&\makecell[l]{$47$}&\makecell[l]{$6$}&\makecell[l]{$66$}&\makecell[l]{$32$}&\makecell[l]{$56$}&\makecell[l]{$48$}\\
 \hline
 TLMSE&\makecell[l]{61}&\makecell[l]{$\mathbf{86}$}&\makecell[l]{$0$}&\makecell[l]{$81$}&\makecell[l]{50}&\makecell[l]{28}&\makecell[l]{-62}&\makecell[l]{1}&\makecell[l]{14}&\makecell[l]{$17$}&\makecell[l]{$27.6$}\\
 \hline
TLMSE+Noise&\makecell[l]{71}&\makecell[l]{83}&\makecell[l]{4}&\makecell[l]{85}&\makecell[l]{62}&\makecell[l]{36}&\makecell[l]{-38}&\makecell[l]{68}&\makecell[l]{30}&\makecell[l]{30}&\makecell[l]{$43.1$}\\
\hline
 TLRR&\makecell[l]{71}&\makecell[l]{85}&\makecell[l]{7}&\makecell[l]{83}&\makecell[l]{66}&\makecell[l]{32}&\makecell[l]{-52}&\makecell[l]{$76$}&\makecell[l]{25}&\makecell[l]{$29$}&\makecell[l]{$42.2$}\\
 \hline
TLLAR&\makecell[l]{71}&\makecell[l]{83}&\makecell[l]{10}&\makecell[l]{81}&\makecell[l]{68}&\makecell[l]{31}&\makecell[l]{-60}&\makecell[l]{65}&\makecell[l]{32}&\makecell[l]{$23$}&\makecell[l]{$40.4$}\\
 \hline
TLENR&\makecell[l]{65}&\makecell[l]{82}&\makecell[l]{12}&\makecell[l]{$\mathbf{87}$}&\makecell[l]{63}&\makecell[l]{53}&\makecell[l]{-13}&\makecell[l]{58}&\makecell[l]{31}&\makecell[l]{$16$}&\makecell[l]{$45.4$}\\
\hline
\textbf{HL+WMME(ours)}&\makecell[l]{$\mathbf{83}$}&\makecell[l]{$75$}&\makecell[l]{$\mathbf{29}$}&\makecell[l]{$81$}&\makecell[l]{$\mathbf{82}$}&\makecell[l]{$\mathbf{64}$}&\makecell[l]{$\mathbf{57}$}&\makecell[l]{$\mathbf{81}$}&\makecell[l]{$\mathbf{83}$}&\makecell[l]{$\mathbf{60}$}&\makecell[l]{$\mathbf{69.5}$}\\
\hline
\end{tabular}
\end{table*}

\begin{figure*}
\centering
\subfloat[]{\includegraphics[height=1.5in,width=2.25in]{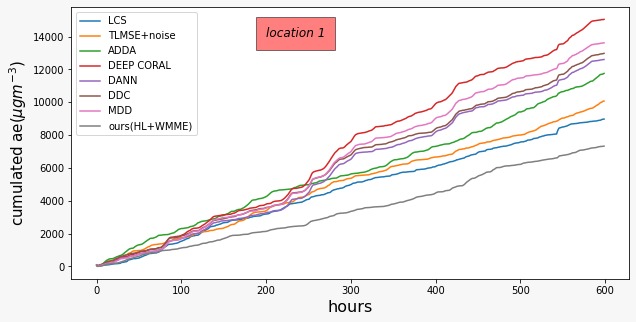}}
\subfloat[]{\includegraphics[height=1.5in,width=2.25in]{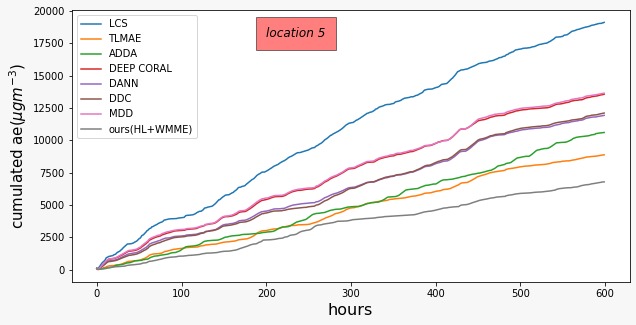}}
\subfloat[]{\includegraphics[height=1.5in,width=2.25in]{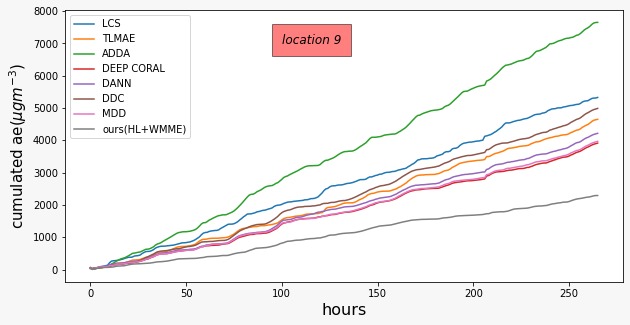}}
\vspace{-1mm}
\caption{Performance evaluation in terms of Cumulated absolute error ($\mu$gm$^{-3}$), accumulated over deployment time at target locations 1, 5 and 9 respectively.}
\label{fig:cumulative ae}
\vspace{-5mm}
\end{figure*}
\begin{figure*}
\centering
\subfloat[]{\includegraphics[height=1.5in,width=2.25in]{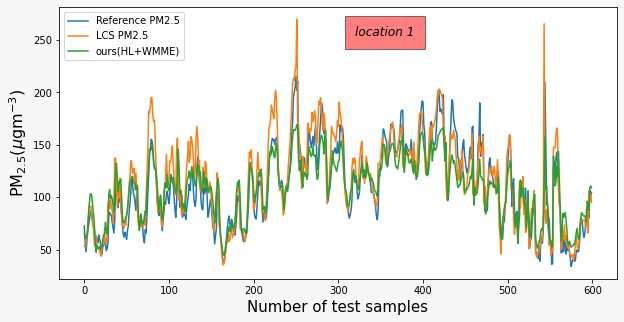}}
\subfloat[]{\includegraphics[height=1.5in,width=2.25in]{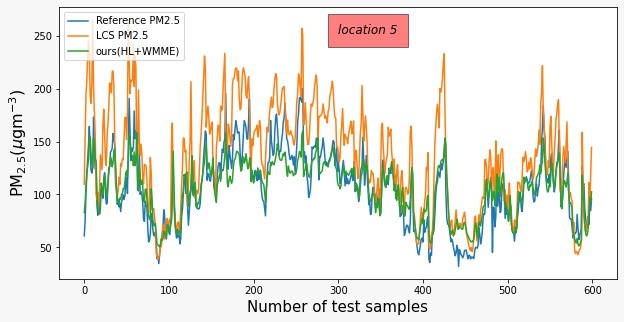}}
\subfloat[]{\includegraphics[height=1.5in,width=2.25in]{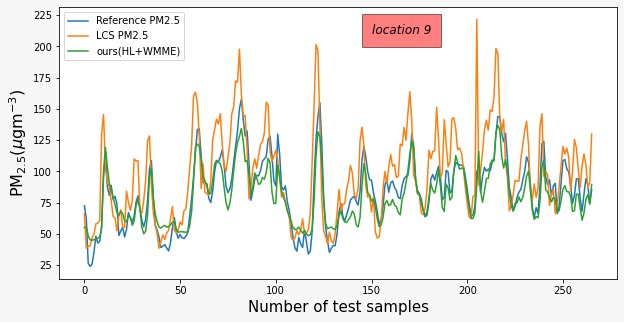}}
\vspace{-1mm}
\caption{Depicting the PM$_{2.5}$ values over time at target location 1, 5, 9 respectively measured by reference monitor and the corresponding uncalibrated PM$_{2.5}$ values measured by low-cost sensors, calibrated PM$_{2.5}$ values by our model.}
\label{fig:predicted PM2.5}
\vspace{-5mm}
\end{figure*}
\begin{table*}[t]
\centering
\caption{MAE (standard deviation) at different target locations for different models. Location indices are in columns and models are in rows. The smaller the better.}
\label{table:mae}
\begin{tabular}{|p{2.1cm}||p{0.9cm}|p{0.9cm}|p{0.9cm}|p{0.9cm}|p{0.9cm}|p{0.9cm}|p{0.9cm}|p{0.9cm}|p{0.9cm}|p{0.9cm}|p{0.9cm}|}
 \hline
 \multicolumn{12}{|c|}{MAE (std. dev.)}\\
 \hline
   \textbf{Model} & loc1 & loc2 & loc3 & loc4 & loc5 & loc6 & loc7 & loc8 & loc9 & loc10 & avg. \\ [0.5ex]
 \hline
  UNCAL&\makecell[l]{14.98\\(14.47)}&\makecell[l]{$15.91$\\(13.21)}&\makecell[l]{49.41\\(37.72)}&\makecell[l]{$23.00$\\(19.88)}&\makecell[l]{31.98\\(22.25)}&\makecell[l]{23.41\\(16.89)}&\makecell[l]{41.62\\(41.29)}&\makecell[l]{$12.87$\\(9.98)}&\makecell[l]{$20.20$\\(16.68)}&\makecell[l]{20.78\\(16.91)}&\makecell[l]{29.54}\\
 \hline
 DANN&\makecell[l]{21.02\\(19.64)}&\makecell[l]{12.29\\(11.12)}&\makecell[l]{37.30\\(32.12)}&\makecell[l]{15.28\\(12.18)}&\makecell[l]{19.91\\(14.67)}&\makecell[l]{17.58\\(13.53)}&\makecell[l]{11.39\\(12.06)}&\makecell[l]{17.00\\(12.42)}&\makecell[l]{15.88\\(11.56)}&\makecell[l]{30.34\\(19.77)}&\makecell[l]{19.80}\\
 \hline
 ADDA&\makecell[l]{19.65\\(14.07)}&\makecell[l]{33.02\\(20.18)}&\makecell[l]{26.44\\(18.65)}&\makecell[l]{42.96\\(26.55)}&\makecell[l]{17.72\\(14.24)}&\makecell[l]{36.65\\(23.20)}&\makecell[l]{22.81\\(13.40)}&\makecell[l]{18.25\\(12.70)}&\makecell[l]{28.78\\(18.61)}&\makecell[l]{38.32\\(25.98)}&\makecell[l]{28.46}\\
  \hline
 DEEP-CORAL&\makecell[l]{25.09\\(19.36)}&\makecell[l]{11.60\\(10.28)}&\makecell[l]{36.92\\(31.53)}&\makecell[l]{22.32\\(18.52)}&\makecell[l]{22.65\\(16.84)}&\makecell[l]{18.18\\(13.90)}&\makecell[l]{12.45\\(12.87)}&\makecell[l]{21.83\\(15.33)}&\makecell[l]{14.78\\(11.03)}&\makecell[l]{30.77\\(19.73)}&\makecell[l]{21.66}\\
   \hline
  DDC&\makecell[l]{21.64\\(18.89)}&\makecell[l]{13.01\\(10.32)}&\makecell[l]{40.46\\(35.49)}&\makecell[l]{20.58\\(16.25)}&\makecell[l]{20.21\\(15.73)}&\makecell[l]{25.79\\(13.64)}&\makecell[l]{38.28\\(12.76)}&\makecell[l]{25.96\\(14.56)}&\makecell[l]{18.80\\(11.15)}&\makecell[l]{30.62\\(21.99)}&\makecell[l]{25.54}\\
  \hline
  MDD&\makecell[l]{22.71\\(19.90)}&\makecell[l]{11.59\\(10.48)}&\makecell[l]{37.78\\(29.07)}&\makecell[l]{14.76\\(13.41)}&\makecell[l]{22.79\\(17.29)}&\makecell[l]{17.30\\(14.26)}&\makecell[l]{10.88\\(10.25)}&\makecell[l]{16.90\\(12.26)}&\makecell[l]{14.95\\(11.07)}&\makecell[l]{31.74\\(19.74)}&\makecell[l]{20.14}\\
  \hline
 TLMAE&\makecell[l]{19.14\\(14.78)}&\makecell[l]{14.57\\(13.73)}&\makecell[l]{28.33\\(28.93)}&\makecell[l]{17.86\\(14.40)}&\makecell[l]{14.81\\(11.21)}&\makecell[l]{17.77\\(12.36)}&\makecell[l]{11.75\\(9.12)}&\makecell[l]{15.63\\(11.98)}&\makecell[l]{17.61\\(14.21)}&\makecell[l]{20.85\\(14.49)}&\makecell[l]{17.83}\\
 \hline
 TLMSE&\makecell[l]{19.15\\(14.83)}&\makecell[l]{10.64\\(8.39)}&\makecell[l]{27.81\\(27.82)}&\makecell[l]{17.49\\(14.01)}&\makecell[l]{19.99\\(14.27)}&\makecell[l]{20.44\\(14.66)}&\makecell[l]{16.19\\(10.94)}&\makecell[l]{20.43\\(18.87)}&\makecell[l]{24.95\\(17.07)}&\makecell[l]{28.75\\(19.97)}&\makecell[l]{20.58\\}\\
 \hline
TLMSE+Noise&\makecell[l]{16.81\\(12.49)}&\makecell[l]{11.62\\(9.59)}&\makecell[l]{27.30\\(20.10)}&\makecell[l]{15.83\\(12.13)}&\makecell[l]{17.88\\(12.08)}&\makecell[l]{19.64\\(13.19)}&\makecell[l]{14.51\\(10.68)}&\makecell[l]{15.46\\(11.24)}&\makecell[l]{17.66\\(13.80)}&\makecell[l]{26.56\\(18.06)}&\makecell[l]{18.33\\}\\
\hline
 TLRR&\makecell[l]{16.73\\(12.47)}&\makecell[l]{10.90\\(9.03)}&\makecell[l]{26.14\\(27.41)}&\makecell[l]{16.96\\(13.51)}&\makecell[l]{16.23\\(12.22)}&\makecell[l]{20.50\\(13.27)}&\makecell[l]{15.23\\(11.25)}&\makecell[l]{13.48\\(11.62)}&\makecell[l]{18.02\\(14.47)}&\makecell[l]{26.71\\(18.37)}&\makecell[l]{18.09\\}\\
 \hline
TLLAR&\makecell[l]{16.82\\(12.49)}&\makecell[l]{11.74\\(9.50)}&\makecell[l]{26.45\\(25.91)}&\makecell[l]{16.60\\(14.74)}&\makecell[l]{15.75\\(11.64)}&\makecell[l]{20.62\\(13.35)}&\makecell[l]{15.46\\(11.74)}&\makecell[l]{16.33\\(9.61)}&\makecell[l]{17.27\\(13.66)}&\makecell[l]{27.60\\(19.10)}&\makecell[l]{18.46\\}\\
 \hline
TLENR&\makecell[l]{18.28\\(12.30)}&\makecell[l]{12.00\\(9.80)}&\makecell[l]{26.22\\(26.25)}&\makecell[l]{14.25\\(12.63)}&\makecell[l]{17.51\\(11.87)}&\makecell[l]{16.99\\(10.95)}&\makecell[l]{12.68\\(10.25)}&\makecell[l]{17.78\\(12.90)}&\makecell[l]{17.06\\95.92)}&\makecell[l]{29.29\\(19.45)}&\makecell[l]{18.21\\}\\
\hline
\textbf{HL+WMME(ours)}&\makecell[l]{$\mathbf{12.22}$\\(9.73)}&\makecell[l]{13.22\\(12.61)}&\makecell[l]{$\mathbf{24.03}$\\(15.49)}&\makecell[l]{15.31\\(15.92)}&\makecell[l]{$\mathbf{11.31}$\\(9.33)}&\makecell[l]{$\mathbf{13.29}$\\(11.47)}&\makecell[l]{$\mathbf{7.37}$\\(6.09)}&\makecell[l]{$\mathbf{12.28}$\\(9.20)}&\makecell[l]{$\mathbf{8.64}$\\(6.09)}&\makecell[l]{$\mathbf{18.29}$\\(15.39)}&\makecell[l]{$\mathbf{13.60}$\\}\\
\hline
\end{tabular}
\end{table*}
\subsection{Implementation Details}
The architecture for our employed deep neural network (DNN) in the proposed scheme is  $512-256-256-256-256-200-K$, i.e., 6 hidden layers and one output layer. The output layer corresponds to the histogram bins, with $K$ bins. All hidden layers have ReLu activation, while the output layer has softmax activation. We train our model iteratively for 200 epochs at each target location using Adam optimizer \cite{kingma2014adam} with an initial learning rate of $10^{-3}$. The target distribution $p(y|\mathbf{x})$ for the histogram loss, eq.~\eqref{eq:histloss}, is chosen as a truncated Gaussian with mean as the ground truth $y$ and a fixed standard deviation equal to the square root of the width of a bin. 
During training, the support of the histogram is chosen as $y\in[0,800]$ as found from the labels in both source and target domains.

The hyperparameter values in the proposed method are set using grid search.
To determine $K$, we perform a grid search starting from $20$ bins to $1220$ bins with a step size of $40$ bins. The optimal $K$ is chosen based on the best R$^2$ score obtained on a target location's validation set at the end of training. 
The value of $|\alpha|_\infty$ is chosen by grid search at $\{0.1,1\}$, and the optimal value is decided by the performance of model on validation set. Only at last location $|\alpha|_\infty$ is chosen from $\{0.001,0.01,0.1,1\}$. At every target location, $T_1, T_2$ are chosen as $15, 80$, respectively, and $\beta = 1$.

\subsection{Baselines}
We compare our model with several important regression methods as baselines. Some of these models are derived from unsupervised domain adaptation (UDA) methods for classification, while some are derived from supervised domain adaptation approaches based on transfer learning. The UDA methods for classification are adapted to regression task by changing their loss to mean-squared error loss.
They have also been changed to semi-supervised by including a few labeled data from target domain, and adding a corresponding supervised loss term for training.

Below we list all the baseline approaches used in the experiments:
\begin{itemize}
    \item UNCAL This provides raw uncalibrated PM$_{2.5}$ values from low-cost sensors.
    \item DANN\cite{ganin2016domain} uses a domain classifier to match source and target distributions. Though this method is one of the popular existing UDA methods, we have modified this to semi-supervised. As we are performing a regression task, the cross-entropy loss is changed to mean-squared error loss for labeled data from both the domains.
    \item ADDA\cite{tzeng2017adversarial} is the same as DANN\cite{ganin2016domain} with the only difference being that domain adaptation is performed in $3$ steps. First, the model learns the source encoder in a supervised way using the labeled data from the source domain. In the second step, it learns the target encoder through adversarial training over the unlabeled target data. In last step, the final model (encoder + regressor) is fine tuned with some labeled data from the target domain.
    \item DEEP-CORAL \cite{sun2016deep}: To alleviate the problem of overfitting to the source distribution, it learns a new feature representation by minimizing the norm of the difference between co-variance matrices of encoded source and target data. 
    \item DDC \cite{tzeng2014deep} is as same as DEEP-CORAL with the only difference being that it places an adaptation layer in the model which learns a new domain-invariant feature representation by minimizing the norm of difference between the mean of encoded source and target data.
    \item MDD \cite{zhang2019bridging} learns a new domain-invariant feature representation for classification by minimizing the margin disparity discrepancy between the encoded feature spaces of source and target domains. To tailor it to a regression task, we calculate the disparity between the regressor and the auxiliary regressor through mean absolute error.
    \item TLMAE and TLMSE \cite{jha2021domain} apply simple transfer learning for their regressors trained to optimize mean absolute error and mean squared error, respectively. These are supervised methods. They learn source encoder with the labeled data at source location and are fine tuned at various target locations using the little labeled training data.
    \item TLRR, TLLAR and TLENR \cite{tibshirani1996regression, li2014forecasting}: These are the same as TLMSE except that they use a regularization term along with the mean squared error loss. TLRR uses L$_2$ regularization, TLLAR uses L$_1$ regularization, and TLENR uses both L$_1$ and L$_2$ regularization.
    \item TLMSE+Noise is the same as TLMSE except that it augments white noise with variance $10^{-2}$ in the ground truths of labeled source and target data. 
\end{itemize}
\subsection{Evaluation Metrics}
The performance of our model is determined by comparing the calibrated values of PM$_{2.5}$ against PM$_{2.5}$ values from reference monitors. The metrics used for this comparison are mean absolute error (MAE) and co-efficient of determination (R$^2$ score). We also compute standard deviation corresponding to the MAE.  
\begin{table*}[t]
\centering
\caption{Ablation analysis: MAE at different target locations. The location indices are in columns and different ablated models are in rows. The smaller the MAE, the better is the performance.}
\label{table:mae_ablation}
\begin{tabular}{|p{2.1cm}||p{0.9cm}|p{0.9cm}|p{0.9cm}|p{0.9cm}|p{0.9cm}|p{0.9cm}|p{0.9cm}|p{0.9cm}|p{0.9cm}|p{0.9cm}|p{0.9cm}|}
 \hline
 \multicolumn{12}{|c|}{MAE}\\
 \hline
   \textbf{Model} & loc1 & loc2 & loc3 & loc4 & loc5 & loc6 & loc7 & loc8 & loc9 & loc10 & avg. \\ [0.5ex]
  \hline
 HL+MME&\makecell[l]{15.71}&\makecell[l]{15.79}&\makecell[l]{$\mathbf{22.91}$}&\makecell[l]{24.06}&\makecell[l]{13.92}&\makecell[l]{15.53}&\makecell[l]{8.12}&\makecell[l]{17.93}&\makecell[l]{10.48}&\makecell[l]{19.25}&\makecell[l]{16.37}\\
 \hline
 HL &\makecell[l]{15.45}&\makecell[l]{15.04}&\makecell[l]{32.58}&\makecell[l]{16.88}&\makecell[l]{13}&\makecell[l]{15.88}&\makecell[l]{8.48}&\makecell[l]{14.47}&\makecell[l]{9.81}&\makecell[l]{18.33}&\makecell[l]{15.99}\\
\hline
 HL+WME&\makecell[l]{12.82}&\makecell[l]{17.25}&\makecell[l]{23.42}&\makecell[l]{15.90}&\makecell[l]{12.47}&\makecell[l]{$\mathbf{13.22}$}&\makecell[l]{7.78}&\makecell[l]{15.30}&\makecell[l]{9.50}&\makecell[l]{18.33}&\makecell[l]{14.60}\\
 \hline
 HL(DD)+WMME&\makecell[l]{20.91}&\makecell[l]{16.19}&\makecell[l]{24.64}&\makecell[l]{19.74}&\makecell[l]{13.88}&\makecell[l]{14.28}&\makecell[l]{9.70}&\makecell[l]{19.89}&\makecell[l]{9.26}&\makecell[l]{22.30}&\makecell[l]{17.08}\\
 \hline
\textbf{HL+WMME}&\makecell[l]{$\mathbf{12.22}$}&\makecell[l]{$\mathbf{13.22}$}&\makecell[l]{24.03}&\makecell[l]{$\mathbf{15.31}$}&\makecell[l]{$\mathbf{11.31}$}&\makecell[l]{13.29}&\makecell[l]{$\mathbf{7.37}$}&\makecell[l]{$\mathbf{12.28}$}&\makecell[l]{$\mathbf{8.64}$}&\makecell[l]{$\mathbf{18.29}$}&\makecell[l]{$\mathbf{13.60}$\\}\\
\hline
\end{tabular}

\end{table*}
\begin{table*}[t]
\centering
\caption{Ablation analysis: R$^2$ scores at different target locations. The location names are in columns and different ablated models are in rows.}
\label{table:r2_score_ablation}
\begin{tabular}{|p{2.1cm}||p{0.9cm}|p{0.9cm}|p{0.9cm}|p{0.9cm}|p{0.9cm}|p{0.9cm}|p{0.9cm}|p{0.9cm}|p{0.9cm}|p{0.9cm}|p{0.9cm}|}
 \hline
 \multicolumn{12}{|c|}{R$^2$ score}\\
 \hline
   \textbf{Model} & loc1 & loc2 & loc3 & loc4 & loc5 & loc6 & loc7 & loc8 & loc9 & loc10 & avg. \\ [0.5ex]
  \hline
 HL+MME&\makecell[l]{68}&\makecell[l]{$64$}&\makecell[l]{26}&\makecell[l]{60}&\makecell[l]{72}&\makecell[l]{50}&\makecell[l]{48}&\makecell[l]{$48$}&\makecell[l]{$74$}&\makecell[l]{55}&\makecell[l]{56.5}\\
 \hline
 HL &\makecell[l]{76}&\makecell[l]{66}&\makecell[l]{-58}&\makecell[l]{79}&\makecell[l]{78}&\makecell[l]{54}&\makecell[l]{42}&\makecell[l]{70}&\makecell[l]{77}&\makecell[l]{57}&\makecell[l]{51.7}\\
\hline
 HL+WME&\makecell[l]{82}&\makecell[l]{$58$}&\makecell[l]{25}&\makecell[l]{79}&\makecell[l]{79}&\makecell[l]{64}&\makecell[l]{49}&\makecell[l]{$64$}&\makecell[l]{$78$}&\makecell[l]{58}&\makecell[l]{63.6}\\
 \hline
 HL(DD)+WMME&\makecell[l]{49}&\makecell[l]{$64$}&\makecell[l]{13}&\makecell[l]{75}&\makecell[l]{74}&\makecell[l]{58}&\makecell[l]{35}&\makecell[l]{$39$}&\makecell[l]{$73$}&\makecell[l]{45}&\makecell[l]{52.5}\\
 \hline
\textbf{HL+WMME}&\makecell[l]{$\mathbf{83}$}&\makecell[l]{$\mathbf{75}$}&\makecell[l]{$\mathbf{29}$}&\makecell[l]{$\mathbf{81}$}&\makecell[l]{$\mathbf{82}$}&\makecell[l]{$\mathbf{64}$}&\makecell[l]{$\mathbf{57}$}&\makecell[l]{$\mathbf{81}$}&\makecell[l]{$\mathbf{83}$}&\makecell[l]{$\mathbf{60}$}&\makecell[l]{$\mathbf{69.5}$}\\
\hline
\end{tabular}

\end{table*}

\subsection{Experimantal Results}
Tables~\ref{table:r2_score} and \ref{table:mae}, show the results of different models, the baselines and the proposed, at $10$ target locations. Our proposed method outperforms the baselines in terms of R$^2$ score and MAE values at all but two locations. 
Moreover, on an average, the proposed method is better than the baselines.
The average R$^2$ score of the proposed method is $21.5$ units higher than the baseline with the best average R$^2$ score, and the average MAE of the proposed method is 4.23 units lower than that of the best baseline.
The results are pictorially depicted at $3$ different target locations in Fig.~\ref{fig:cumulative ae} in terms of the absolute error accumulated over time.
The improvement in the quality of the calibrated signal can be seen in Figure~\ref{fig:predicted PM2.5}. The calibrated signal matches the reference one more reliably as compared to the raw LCS signal.
These results indicate the efficacy of the proposed calibration method in making the measurements more reliable.
\subsection{Ablation analysis}
To analyse the contribution of each component of the proposed method, we perform an ablation analysis, i.e., assessing the performance of the model without the respective component. The results with the ablation analysis are reported in Tables~\ref{table:mae_ablation} and \ref{table:r2_score_ablation}. 
The last rows in both these tables contain results for the un-ablated model, named HL+WMME, while the rows above contain those for different ablated models.

\subsubsection{Removal of the weighting mechanism} If instead of weighted min-max entropy loss (eq.~\ref{eq:wentloss}), we use the unweighted min-max entropy loss (eq.~\ref{eq:entloss}, the performance degrades. We call the former as HL+WMME and the latter as HL+MME in Tables~\ref{table:mae_ablation} and \ref{table:r2_score_ablation}. 


\subsubsection{Only supervised training} Does the unsupervised training give a leverage to the performance? To analyse this, we train the model with only labeled data from both source and target locations. This model is named as HL. The metrics in Tables~\ref{table:mae_ablation} and \ref{table:r2_score_ablation} show a clear improvement with HL+WMME model as compared to the performance with HL only.

\subsubsection{Changing the min-max to only minimization of entropy}
If instead of the min-max entropy, we only minimize the entropy over the unlabeled data with respect to both $\theta$ and $\phi$, the results degrade. We call this model as HL+WME (weighted minimization of entropy) in Tables~\ref{table:mae_ablation} and \ref{table:r2_score_ablation}. Although, minimizing entropy over unlabeled data may appear more useful but the min-max actually helps in making $F_\theta$ generalize better across domains.


\subsubsection{Changing $p(y|\mathbf{x})$ from truncated Gaussian to Dirac delta} If we change the target distribution in the histogram loss to Dirac delta distribution, it takes the form of categorical cross entropy that is popularly used for multi-class classification. However, we see a degradation in performance metrics with this. In Tables~\ref{table:mae_ablation} and \ref{table:r2_score_ablation}, we call this model as HL(DD)+WMME. 

In general, we notice that increasing the model variance helps in generalizing better, and thereby, leads to a better adaptation across domains. This is done in this work by maximization of entropy and by choosing truncated Gaussian distribution as target.
Removing any of these two components, degrades the model performance.

    
\section{Conclusion}
The paper proposes a semi-supervised domain adaptation method for regression and applies it to sensor calibration for air quality monitoring. A major benefit of the proposed approach for the sensor calibration application is its ability to leverage unlabeled data. 
The method achieves better performance as compared to those using only labeled data. Moreover, it also outperforms several existing semi-supervised domain adaptation methods applied to this task. As future work, we would like to apply the proposed method to other regression problems. We would also like to study the application of the proposed method to tackle the problem of sensor drift, i.e., the change of sensor characteristics with time.

\section*{Acknowledgments}
The data used for this work has been collected with the generous support of Maharashtra pollution control board and Bloomberg Philanthropies. SD has been supported by Prasar Bharati.



\bibliographystyle{IEEEtran}
\bibliography{sensor_citation.bib}

\newpage

 




\vfill

\end{document}